\documentclass[sigconf]{acmart}

\usepackage{multirow}
\usepackage{makecell}
\usepackage{amsmath}
\usepackage{amsthm}
\usepackage{booktabs}
\usepackage{tabularx}
\usepackage{color}
\usepackage{xcolor}
\usepackage{algorithm}
\usepackage[noend]{algpseudocode}
\usepackage{subcaption}
\usepackage{float}
\usepackage{url}
\usepackage{graphicx}

\newcommand{\ie}{\textit{i.e.}}

\newtheorem{assumption}{Assumption}

\AtBeginDocument{%
  }

\setcopyright{acmlicensed}
\copyrightyear{2018}
\acmYear{2018}
\acmDOI{XXXXXXX.XXXXXXX}
\acmConference[Conference acronym 'XX]{}{June 03--05,
  2018}{Woodstock, NY}
  
\begin{document}

\title{Toward Robust Semi-supervised Regression via Dual-stream Knowledge Distillation}

\author{Ye Su}
\email{suye@cigit.ac.cn}

\affiliation{%
 \institution{Chongqing Institute of Green and Intelligent Technology, Chinese Academy of Sciences}
 \institution{Chongqing School, University of Chinese Academy of Sciences}
  \country{Chongqing, China}
}
\author{Hezhe Qiao}
\authornotemark[1]
\email{hezheqiao.2022@phdcs.smu.edu.sg}
\affiliation{%
 \institution{School of Computing and Information
Systems\\ Singapore Management University}
  \country{Singapore}
}

 \author{Wei Huang}
 \email{huangwei@bupt.edu.cn}
\affiliation{%
 \institution{Beijing University of Posts and Telecommunications}
   \country{Beijing, China}
}

 \author{Hui He}
 \email{huihe@smu.edu.sg}
\affiliation{%
 \institution{Singapore Management University}
   \country{Singapore}
}

 \author{Lin Chen}
 \email{chenlin@cigit.ac.cn}
\authornote{Corresponding author}
\affiliation{%
 \institution{Chongqing Institute of Green and Intelligent Technology, Chinese Academy of Sciences}
   \country{Chongqing, China}
}

\renewcommand{\shortauthors}{Su et al.}

\begin{abstract}
Semi-supervised regression (SSR), which aims to predict continuous scores of samples while reducing reliance on a large amount of labeled data, has recently received considerable attention across various applications, including computer vision, natural language processing, and audio and medical analysis. Existing SSR methods typically train models on scarce labeled data by introducing constraint-based regularization or ordinal ranking to reduce overfitting.  However, these approaches fail to fully exploit the abundance of unlabeled samples. While consistency-driven pseudo-labeling methods attempt to incorporate unlabeled data, they are highly sensitive to pseudo-label quality and noisy predictions.
To address these challenges, we introduce a  
\textbf{D}ual-stream  \textbf{K}nowledge \textbf{D}istillation framework (\textbf{DKD}), which is specially designed for the SSR task to distill knowledge from both continuous-valued knowledge and distribution information, better preserving regression magnitude information and improving sample efficiency. Specifically, in DKD, the teacher is optimized solely with ground-truth labels for label distribution estimation, while the student learns from a mixture of real labels and teacher-generated pseudo targets on unlabeled data. The distillation design ensures the effective supervision transfer, allowing the student to leverage pseudo labels more robustly. Then, we introduce an advanced Decoupled Distribution Alignment (DDA) module, which separately aligns the target and non-target distributions between the teacher and student. To improve the reliability of non-target knowledge transfer, DDA incorporates a variance-guided non-target distribution alignment strategy that adaptively downweights uncertain teacher predictions, thereby enhancing the student’s capacity to mitigate noise in pseudo-label supervision and learn a better-calibrated regression predictor.
Extensive experiments were conducted on diverse datasets, including audio, text, image, and medical data. We show that DKD exhibits strong generalization capabilities and outperforms the best competing methods by 4.97\% and  11.81\% in terms of overall MAE and \(R^2\), respectively.  


\end{abstract}



\maketitle

\section{Introduction}

Semi-supervised regression (SSR) aims to rank or predict continuous scores of samples while reducing reliance on a large amount of labeled data. It has been widely applied in various scenarios, including age estimation, clinical score prediction, and audio-based assessments, and other real-world applications where a large amount of labeled data across diverse are difficult to obtain
\cite{ren2022balanced,gui2024survey,bao2024robust}. With the rapid advancement of deep learning in recent years, regression tasks using deep learning models across various domains have attracted significant attention~\cite{yang2024robust,van2020survey}, especially in the SSR task. Some approaches directly employ deep learning models such as CNNs for images and RNNs for time series, optimizing them with loss functions like Mean Squared Error (MSE) or  Mean Absolute Error (MAE) on the available labeled data \cite{mohammadi2024deep,yang2022survey,zhang2017mpiigaze}.
However, such a straightforward application struggles to learn effective patterns in scenarios with scarce labeled data, owing to the heavy reliance of deep learning models on large-scale annotations \cite{pan2024pseudo,yin2022fishermatch}. Therefore, direct regression (DR) based models typically yield suboptimal performance.

To exploit the abundance of unlabeled data, several consistency-driven pseudo-labeling methods that assign the pseudo-labels on the unlabeled samples have been proposed \cite{zhang2025semi,zheng2025language,min2024leveraging}. Pseudo-labels are typically generated by training on the available labeled samples, and consistency regularization is often used to assist the optimization process and pseudo-labeling \cite{zhao2025few,zhao2025variable}. 
While pseudo-labeling helps mitigate the challenge of limited labeled data, as some predicted incorrect labels are directly treated as true labels during training, and the model is likely to overfit to the wrong annotations. Therefore, the effectiveness of these approaches heavily relies on the quality of the pseudo-labels \cite{jo2024deep,yin2022fishermatch}, and assessing the reliability of the generated pseudo-labels remains a challenging task.

\begin{figure}[!t]
  \centering
  \includegraphics[width=1.0\linewidth]{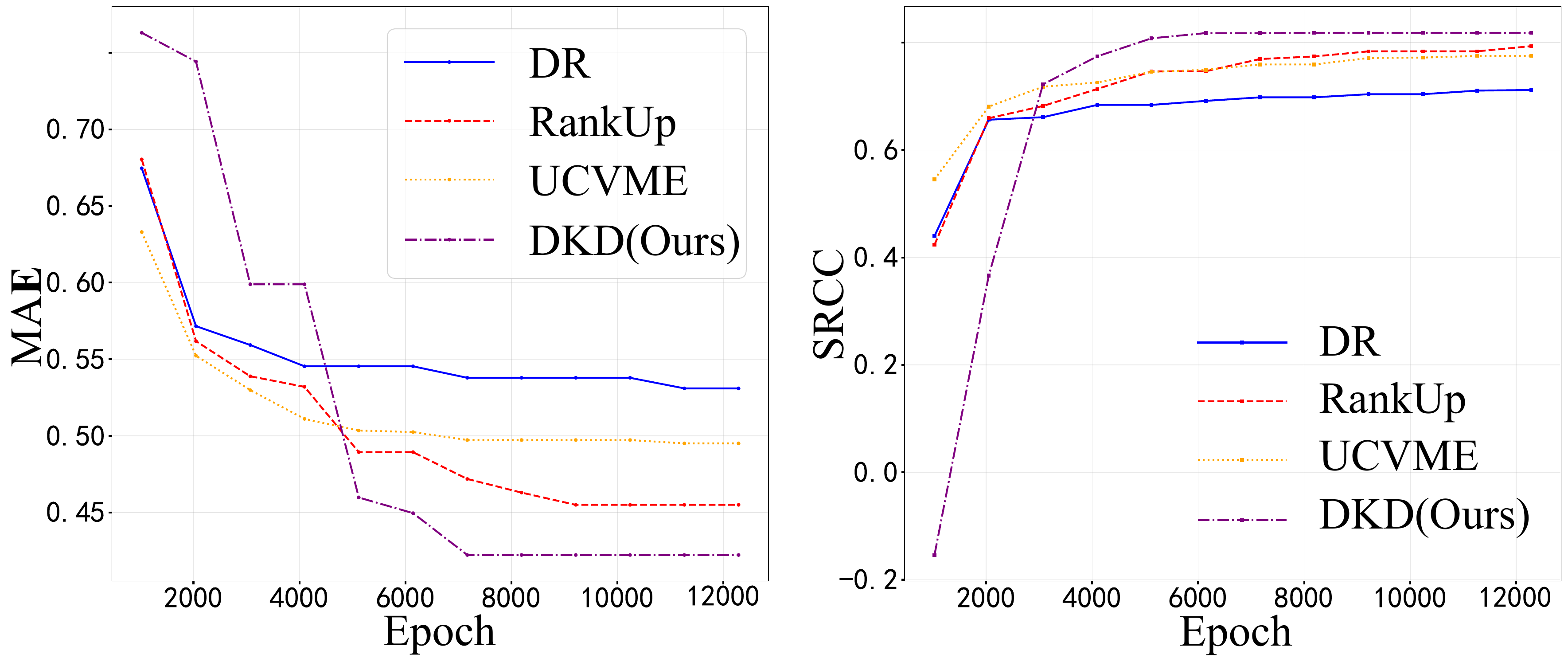}
  \caption{MAE and test SRCC curves for direct regression (DR), Rankup \cite{huang2024rankup}, UCVME \cite{dai2023semi}, and DKD on the BVCC dataset \cite{cooper2021voices}. RankUp achieves the least competitive results, suggesting that its pairwise ranking strategy may not adequately capture continuous relationships. Although UCVME improves performance through consistency-constrained pseudo-labeling, it still underperforms DKD in terms of both MAE and SRCC.} 
  \label{fig:pre}
  \vspace{-2em}
\end{figure}
To address these challenges, we introduce a \textbf{D}ual-stream \textbf{K}nowledge \textbf{D}istillation framework   (\textbf{DKD}), specifically designed for SSR, to effectively exploit unlabeled samples. DKD distills the knowledge from both continuous-valued knowledge and distribution information in an end-to-end manner, better preserving regression magnitude information and improving sample efficiency. Specifically, DKD is implemented by a teacher model optimized solely with ground-truth labels for label distribution learning, producing informative predictions, while the student learns from a mixture of real labels and teacher-generated pseudo scores on unlabeled data, ensuring effective supervision transfer. The distillation allows the student to leverage pseudo labels more robustly, mitigating the interference from noisy pseudo-supervision in SSR.  We further introduce an advanced \textbf{D}ecoupling \textbf{D}istribution \textbf{A}lignment (\textbf{DDA}) approach to distill the distribution information from the labeled trained teacher model, where we decouple the distribution into two parts for alignment, rather than aligning the distribution directly. DDA aligns the target and non-target distributions between the teacher and student in a decoupled where we design a variance-guided non-target distribution alignment strategy that adaptively reduces the influence of uncertain teacher predictions during distillation.
With an adaptive learnable weighting scheme, DDA enables the student to better exploit unlabeled samples while reducing overfitting to noisy pseudo-labels and learning a more well-calibrated regression model. As shown in the Fig \ref{fig:pre}, the DKD can learn more robust and more generalizable knowledge representation provided by the mixture of labeled and pseudo-labeled samples provided by the teacher model. It outperforms the conventional method, direct regression (DR), and strong competing methods, including RankUp \cite{huang2024rankup}, UCVME \cite{dai2023semi}, achieving the best performance on both MAE and SRCC.
In summary, this work makes the following main contributions.

\begin{itemize}
\item  We propose DKD, the first end-to-end decoupled distillation framework for semi-supervised regression, which effectively directly distills both continuous-valued knowledge and distribution information in an end-to-end manner, better preserving regression magnitude information and improving sample efficiency under limited labels.
\item We develop a Decoupled Distribution Alignment (DDA) module for reliable regression distillation, which disentangles target and non-target knowledge transfer and further employs variance-guided non-target alignment to downweight uncertain teacher predictions, thereby reducing pseudo-label noise and improving student calibration. 
\item We provide a theoretical analysis showing that the variance-aware weight reduces gradient noise in the non-target KL term and yields a tighter student-risk bound than constant weight, giving provable justification for the adaptive scheme.
\item Extensive experiments were conducted on the datasets from various domains, including audio, text, image, and medical data. The experimental results demonstrate that DKD outperforms the state-of-the-art semi-supervised regression methods.




\end{itemize}

\section{Related Work}

\subsection{Semi-supervised Classification.} 
Semi-supervised classification (SSC) aims to improve classification performance by leveraging a small amount of labeled data along with a large amount of unlabeled data by generating the corresponding pseudo labels. 
The pseudo-label-based methods typically depend on the smoothness assumption that neighboring data points in the feature space tend to exhibit similar labels \cite{yang2022survey,Meysurvey,berthelot2019mixmatch,sohn2020fixmatch}.  
The pseudo-label generation approach can effectively enhance the dataset and help the model learn fine-grained boundaries \cite{lee2013pseudo}, for example,  Jiang et al. proposed DLPC \cite{JiangScalable}, a scalable semi-supervised learning framework that jointly propagates soft pseudo-labels via a dynamic graph and corrects them using discriminative regression losses on independent class indicators, effectively addressing unreliable similarity structures on boundary samples. Knowledge distillation is also used for pseudo-generation to refine the knowledge from existing methods, for example, Xie \textit{et al.} present a pseudo-labeling approach based on knowledge distillation, using diverse student training techniques. This involves initially training a teacher model on labeled images to generate pseudo-labels for unlabeled examples \cite{xie2020self}. 

Most of the SSC methods are designed for classification tasks, where only high-confidence predictions are retained for training. However, identifying high-confidence samples is much more challenging in SSR. To address this issue, we propose dual knowledge distillation, which combines continuous-value distillation and DDA to mitigate the negative impact of inaccurate teacher predictions. 
In addition, although some knowledge distillation methods have been applied to SSC,  their full potential has not been explored in the SSR task.

\subsection{Semi-supervised Regression}
The SSR methods can be roughly categorized into pseudo-label-based methods, consistency regularization, and contrastive-learning-based methods.
Jo \textit{et al.} \cite{jo2024deep} propose an SSR framework that uses uncertainty estimation to guide pseudo-labeling, alongside a pseudo-label calibration approach that enhances pseudo-label quality by propagating information from labeled to unlabeled samples.
Semi-supervised deep kernel learning  \cite{jean2018semi} minimizes predictive variance on unlabeled data through consistency regularization, leveraging the robust uncertainty quantification provided by Gaussian processes. 
Some contrastive-learning-based methods exploit the ranking information among samples as auxiliary for regression \cite{qiao2022ranking,huang2024rankup}.  For example, RankUp \cite{huang2024rankup} constructs pairwise ranking labels from continuous targets with an auxiliary ranking classifier. CLSS \cite{dai2023semi} and GCLSS \cite{wang2025contrastive} facilitate contrastive regression by exploiting the ordinal relationships between unlabeled samples with a spectral seriation algorithm. 

Although these methods have achieved remarkable success in SSR, most of them do not fully leverage the potential of abundant unlabeled data. Pseudo-labeling approaches incorporate unlabeled samples into training, but they are often sensitive to noisy pseudo-targets and may even degrade performance. In DKD, we propose an advanced decoupled representation distillation strategy that transfers knowledge from both continuous-value supervision and distributional information, enabling effective utilization of both pseudo-labeled and ground-truth labeled samples.

\subsection{Distribution Alignment in Knowledge Distillation}
Distribution alignment is a strategy used during student model training to match not only the outputs of the teacher model but also its underlying distributions. Existing approaches can be roughly categorized into logit-level alignment and feature-level alignment.
Logit-level alignment aims to align the output distribution (e.g., softmax probabilities) of the teacher and student \cite{hinton2015distilling,zhao2022decoupled,zeng2025normality}. For example, the multi-level logit distillation method \cite{jin2023multi} employs multi-level prediction alignment to facilitate instance prediction learning in the student model. Curriculum temperature \cite{li2023curriculum} introduces temperature scheduling for dynamically adjusting softening intensity across training phases, which smooths gradient flow.
The feature-level distribution alignment matches the representation, which is often done by maximum mean discrepancy, contrastive, and correlation-based loss \cite{gou2021knowledge,chen2021wasserstein}. Notable advancements in feature-level distillation encompass DiffKD \cite{huang2023knowledge}, training diffusion denoisers with teacher features to enforce attention map congruence in critical regions; WCoRD \cite{chen2021wasserstein}, constructing global-local hybrid contrastive objectives with Wasserstein dual loss to align deep geometric structures. 
These alignment approaches have demonstrated effectiveness in knowledge transfer.  Although there are some attempts by applying the decoupled distillation on the classification task \cite{zhao2022decoupled,li2023decoupled,feng2024cross}. There is no work on semi-supervised regression that more emphasizes how pseudo-labels are leveraged effectively. Another difference of our method is that we incorporate the adaptive learnable weighting scheme specifically for the non-target alignment 
to further prevent the student from overfitting noisy pseudo-supervision. Meanwhile, the continuous-valued knowledge is incorporated into the distribution alignment on the basis of label distribution learning and alignment, to keep the regression magnitude information. 


\begin{figure*}[!t]
  \centering
  \includegraphics[width=0.85\linewidth]{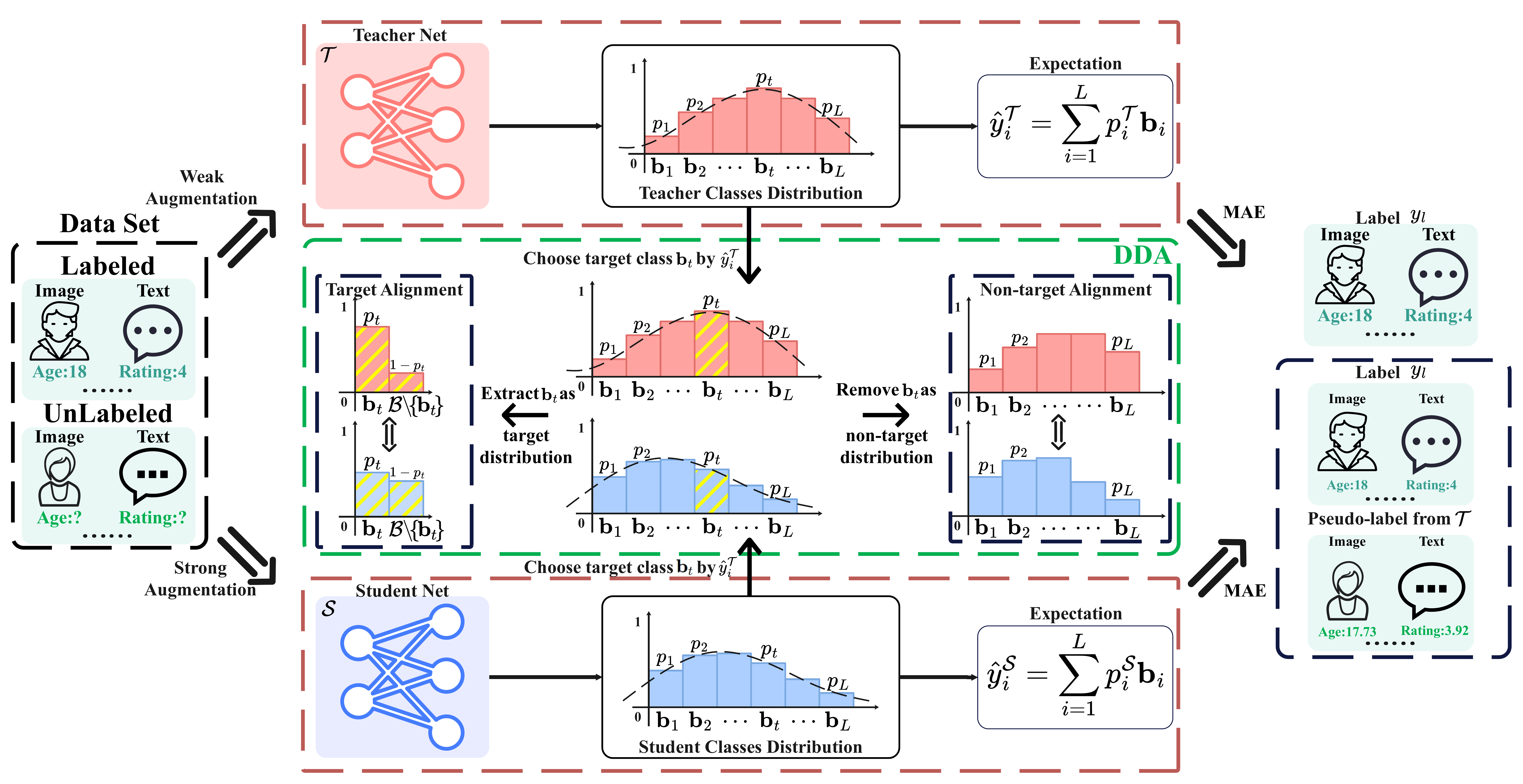}
  \caption{The overview of DKD. (1) The input of DKD includes both
  labeled and unlabeled data. The teacher model is exclusively trained
  on the labeled data and is then used to generate pseudo labels for
  the unlabeled samples. The student model is trained on the entire
  dataset with a mixture of real labels and teacher-generated pseudo
  targets. (2) Both teacher and student are trained under a
  label-distribution formulation of regression, while continuous-valued
  distillation enforces consistency by minimizing the gap between
  their expected scores. (3) In the distribution distillation, DDA
  first identifies the target class to construct a binary probability
  distribution and then aligns the predicted distributions over
  classes between teacher and student separately for the target and
  non-target parts. }
  \label{fig:framework}
\end{figure*}

\section{Methodology}

\noindent\textbf{Notations.} We assume the dataset $\mathcal{X}$
consists of $n$ labeled samples
$\mathcal{X}_l=\{(x_i,y_i)\}_{i=1}^{n}$ and $m$ unlabeled samples
$\mathcal{X}_u=\{x_i\}_{i=n+1}^{n+m}$, with each instance
$x\in\mathbb{R}$ representing image, text, audio, or tabular medical
data. $\mathcal{Y}=\{y_i\}$ denotes the set of ground-truth labels,
where each $y_i\in[0,1]$ is a continuous value standardized into the unit interval from the original label range; the regression magnitude is recovered at evaluation time by inverting the standardization before computing MAE.

\noindent\textbf{Problem Statement.} 
Deep semi-supervised regression can be formulated as follows. 
Given a training set $\mathcal{X}_T = \mathcal{X}_l \cup \mathcal{X}_u$, where $\mathcal{X}_l$ denotes the limited labeled samples and $\mathcal{X}_u$ denotes the unlabeled samples, the goal is to train a mapping function $f: \mathcal{X} \mapsto \mathcal{Y}$. 
The function maps each input sample $x_i \in \mathcal{X}$ to a continuous scalar target $y_i \in \mathbb{R}$. 
Semi-supervised learning explicitly incorporates unlabeled samples to complement the limited labeled data during training, thereby improving the regression model under scarce supervision.

\subsection{The Overview of DKD}
As shown in Figure~\ref{fig:framework}, DKD predicts scores through a
dual-stream distillation framework, where the ground truth is
transformed into a label distribution, effectively converting the
regression task into a discrete distribution estimation problem for
both teacher and student models. We then apply continuous-valued
distillation over the label distribution, and DDA for the distribution
alignment, to enable the student to learn fine-grained and more
generalized knowledge.

This distillation process is particularly effective for SSR due to two
key factors: (1) we reformulate the regression task as joint
distillation on continuous-valued knowledge and distribution
information in an end-to-end manner, which effectively mitigates the
risk of noise when integrating pseudo-labels, especially when labeled
samples are limited; and (2) we introduce a novel decoupled
distribution alignment strategy for distillation over classes,
enabling the model to capture fine-grained features more effectively
than conventional alignment methods.

\subsection{Distillation for Semi-supervised Regression}
\noindent\textbf{Label Distribution Learning.} Conventional SSR
methods typically predict the score by directly mapping the input
sample $x_i$ to the target value $y$. However, this approach is
highly sensitive to noise and irrelevant features, often resulting in
poor generalization and a high risk of overfitting, especially when
few labels are available. Therefore, following
\cite{gao2017deep,xu2019label}, we define $L$ classes
$\mathcal{B}=\{\mathbf{b}_i\}_{i=1}^{L}$ based on the range of
ground-truth $y$, converting the regression task into a discrete
distribution estimation problem. $\mathcal{B}$ can be intuitively
viewed as a set of evenly spaced score bins (anchors) that
discretize the continuous label range into $L$ classes, each
covering an interval of width $c$, with $\mathbf{b}_i=(i-1)\cdot c$.
Uniform discretization is sufficient for our purposes: the
regression magnitude is recovered through the expectation
$\hat y=\sum_i p_i\mathbf b_i$ regardless of the partition, and the
ordinal information among neighbouring bins is preserved by the
distribution itself. We compare uniform binning against
density-based and quantile-based alternatives in
Section~\ref{sec:experiments}.

Then $y$ can be obtained by the expectation
\begin{equation}
 y = \sum_{i=1}^{L}\tilde p_i\,\mathbf{b}_i,
\end{equation}
where $\tilde p_i$ is the model output for class $\mathbf{b}_i$. The
mapping function $f$ therefore learns a distribution
$\mathcal{P}=\{p_i\}_{i=1}^{L}$ over predefined classes. By
transforming numerical regression into an interval-prediction problem,
it offers a smoother learning objective and richer label distribution
than direct numerical
regression~\cite{xu2019label}, leading to more effective use of
limited labels and better noise robustness.

To obtain a probabilistic distribution, we apply the softmax function, and the predicted distribution is formulated as:
\begin{equation}
p_i=\frac{\exp \left(\tilde{p}_i\right)}{\sum_{i=1}^L \exp \left(\tilde{p}_j\right)}
\end{equation} 
where $p_i$ is the predicted probability for the class $\mathbf{b}_i$ after softmax.
These outputs form a distribution, from which we compute the expectation to obtain the final score as follows:
\begin{equation}
\hat y_i = \sum\limits_{i = 1}^L {{p_{i}}} {\mathbf{b}_i}
\end{equation}

Finally, we employ the MAE loss to minimize the distance between the prediction $\hat y_i$ and the ground truth  $y_i$:
\begin{equation}
    \mathcal L = \frac{1}{|{{\mathcal X}}|}\sum_{x_i\in {\mathcal{X}}} |y_i-\hat y_i|
\end{equation}
where $y_i$ and the $\hat{y}_i$ are the ground truth label and the predicted score of sample $x_i$ , respectively. $|{\mathcal X}|$ is the number of samples in ${\mathcal X}$. 
In DKD, both the teacher and student models utilize the MAE loss function to incorporate supervised information, as detailed below. 

\noindent\textbf{Continuous-valued Knowledge Distillation.}
Since label distribution learning alone is insufficient for effective
representation learning, we reformulate the distillation process. We
first apply weak augmentation (e.g., random crop, flip) to the data
before feeding it into the teacher, while strong augmentation (e.g.,
RandAugment~\cite{cubuk2020randaugment}) is applied to the student's
input. The predicted scores from the teacher are used to guide the
student by providing pseudo-labels for the unlabeled samples. The
teacher and student losses are
\begin{equation}
\mathcal{L}_{\mathcal{T}}=\frac{1}{|\mathcal{X}_l|}\sum_{x_i\in\mathcal{X}_l}|y_i-\hat y_i^{\mathcal{T}}|,
\end{equation}
\begin{equation}
\mathcal{L}_{\mathcal{S}}=\frac{1}{|\mathcal{X}|}\sum_{x_i\in\mathcal{X}}|y_i-\hat y_i^{\mathcal{S}}|,
\end{equation}
where $\hat y_i^{\mathcal T}$ and $\hat y_i^{\mathcal S}$ are the
predicted scores of teacher and student, respectively. The overall
DKD objective is
\begin{equation}
\mathcal{L}_{DKD}=\mathcal{L}_{\mathcal T}+\mathcal{L}_{\mathcal S}+\mathcal{L}_{\text{Alignment}}.
\end{equation}

The alignment loss $\mathcal{L}_{\text{Alignment}}$ minimizes the
distance between teacher and student distributions, typically
implemented with KL divergence or MSE. In DKD, we align label
distributions and further introduce a decoupled distribution
alignment strategy, described next.

\subsection{Decoupling Distribution Alignment (DDA)}
Conventional alignment often uses KL divergence or MSE to minimize the
distance between teacher and student. Although distillation helps the
student exploit pseudo-labels more stably, simple alignment may be
misled by noisy targets in pseudo-labeling, yielding suboptimal
performance~\cite{mei2023denkd}. To mitigate this issue, we propose
DDA that aligns the distributions over classes at a more granular
level, improving the student's ability to fully exploit unlabeled
samples.

Inspired by decoupling target-level supervision from non-target
relational insights in classification~\cite{zhao2022decoupled,li2024correlation,Zhengrekd},
we adopt a similar strategy. We first identify the target class to
formulate a binary classification, treating the remaining classes as
non-targets. Alignment is then performed separately on the target and
non-target distributions.

\noindent \textbf{Target class identification.}
Since the entire dataset is fed into the teacher model, it is worth noting that the unlabeled samples do not provide supervision signals—only the labeled samples contribute to the MAE loss in the teacher model. But in the advanced alignment, we choose to involve the unlabeled samples to provide more consistent information. Specifically, to determine the target class, we utilize the ground-truth labels for the labeled samples and the pseudo-labels for the unlabeled ones. 
In the presence of ground-truth labels, the target class $\mathbf{b}_t$ is assigned to the corresponding label $y_i$; otherwise, $\mathbf{b}_t$ can be derived from the teacher's prediction $\hat{y}_i^{\mathcal{T}}$. Furthermore, we denote the probability associated with the target class $\mathbf{b}_t$ as $p_t$. The probability of the non-target set $\mathcal{B} \setminus \{\mathbf{b}_t\}$ is thus defined as $1 - p_t$. 

\noindent \textbf{Variance-guided class distribution alignment.}
Based on the identification of the target class, we can split the class into the target distribution to formulate the $\mathbf{p} = {\{p_t, 1-p_t\}}$, with $\mathbf{p}^{\mathcal{T}}$ and $\mathbf{p^{\mathcal{S}}}$ to represent the teacher and student.  The results of classes refer to the non-target distribution, \ie, the remaining classes after excluding the target class, denoted as  
$\mathbf{q} = \{ p_k \}_{k\neq t}^{L}$,  with $\mathbf{q}^{\mathcal{T}}$ and $\mathbf{q^{\mathcal{S}}}$ to represent the teacher and student. The optimization of DDA can be reformulated as follows:
\begin{equation} \label{eq:weight}
\mathcal{L}_{D D A}=\frac{1}{|{{\mathcal X}}|}\sum_{x_i\in {\mathcal{X}}}   \mathrm{KL}\left(\mathbf{p}_{i}^{\mathcal{T}}, \mathbf{p}_{i}^{\mathcal{S}}\right)+ \beta\cdot\sum_{x_i\in {\mathcal{X}}}\omega_i\cdot \mathrm{KL}\left(  \mathbf{q}_{i}^{\mathcal{T}}, \mathbf{q}_{i}^{\mathcal{S}}\right) 
\end{equation}
where $\mathrm{KL}\left(\mathbf{p}_{i}^{\mathcal{T}} \| \mathbf{p}_{i}^{\mathcal{S}}\right)$ aims to measure the similarity between the teacher distribution and student distribution on the target class, $\mathrm{KL}\left(\mathbf{q}_{i}^{\mathcal{T}} \| \mathbf{q}_{i}^{\mathcal{S}}\right)$ is the similarity on the non-target class.  
$\beta$ is the hyperparameter that controls the weights of no-target class alignment. For each sample, we introduce one learnable weight  $\omega_i= 1 - \sigma_i^{\mathcal{T}}$   where $\sigma^{\mathcal{T}}_i$ is the standard deviation of teacher predictions. The key intuition is that incorporating this variance-aware weighting into the non-target class alignment allows the model to down-weight uncertain samples and focus the alignment on more reliable teacher guidance. As 
target class generally offers more reliable supervision, whereas the non-target part is more susceptible. Therefore, we assign adaptive weights using the variance of the teacher’s predictions as an uncertainty indicator for each sample to downweight uncertain samples and place greater emphasis on reliable teacher guidance.
We apply the weight only to the non-target distribution because the target class generally provides more reliable supervision, whereas the non-target part is more susceptible and therefore benefits more from adaptive weighting.
Please refer to Sec. \ref{sec:var-weight} for a detailed theoretical analysis of this design.

Compared with directly minimizing the KL divergence between teacher and student predictions on a mixture of labeled and pseudo-labeled samples, DDA performs decoupled alignment on target and non-target distributions, and introduces an adaptive learnable weighting scheme specifically for the non-target alignment. This prevents the student from overfitting noisy pseudo-supervision. As a result, DDA provides a more robust distribution alignment, enabling the student to better exploit unlabeled data under mixed supervision of ground-truth labels and pseudo targets.

\subsection{Variance-guided Non-target Weighting: Theoretical
Analysis}\label{sec:var-weight}

We now formalise why the adaptive weight $\omega_i$ in Eq.~\eqref{eq:weight} yields a tighter noise bound than any constant alternative for the non-target KL term. Because $y_i\in[0,1]$ after standardization, $\sigma_i^{\mathcal T}\in[0,1]$ and $\omega_i\in[0,1]$ without further clipping. We present the main result first, then provide the supporting assumption and proof.

Let the gradient of the non-target KL in Eq.~\eqref{eq:weight} be $G$-Lipschitz with respect to the teacher-side distribution.

\begin{theorem}[Gradient-noise reduction]\label{thm:main}
The expected squared norm of the gradient noise satisfies
\begin{equation}
\mathbb{E}\!\left\Vert\omega_i\nabla_{\theta}\mathrm{KL}(\mathbf{q}_i^{\mathcal T}\Vert\mathbf{q}_i^{\mathcal S})
-\omega_i\nabla_{\theta}\mathrm{KL}(\mathbf{q}_i^{\star}\Vert\mathbf{q}_i^{\mathcal S})\right\Vert_2^2
\le (1-\sigma_i^{\mathcal T})^{2}C^{2}G^{2}(\sigma_i^{\mathcal T})^{2},
\label{eq:noise-bound}
\end{equation}
where $G>0$ is the Lipschitz constant. Let $C>0$ be the noise-scale constant bounding the teacher error: $\Vert\boldsymbol{\varepsilon}_i\Vert_2\le C\,\sigma_i^{\mathcal T}$ (formalised in Assumption~\ref{assum:noise}). 
which is strictly smaller than the unit-weight bound $C^2 G^2 (\sigma_i^{\mathcal T})^2$ for every sample with $\sigma_i^{\mathcal T}\in(0,1)$.
\end{theorem}
\begin{theorem}[Tighter risk bound]\label{thm:main1}
Let $\mathcal{R}(\theta)$ denote the population regression risk and $\widehat{\mathcal{L}}^{\omega}_{DDA}$ the empirical DDA loss. Under standard Rademacher complexity conditions, with probability at least $1-\delta$,
\begin{equation}
\mathcal{R}(\theta)\le\widehat{\mathcal{L}}^{\omega}_{DDA}(\theta)
+\underbrace{O\!\left(\sqrt{\tfrac{1}{|\mathcal X|}\textstyle\sum_i
\omega_i^{2}(\sigma_i^{\mathcal T})^{2}}\right)}_{\text{noise term}}
+O\!\left(\sqrt{\tfrac{\log(1/\delta)}{|\mathcal X|}}\right).
\label{eq:risk-bound}
\end{equation}
The noise term is strictly smaller for $\omega_i=1-\sigma_i^{\mathcal T}$ than for any constant weight $\omega= c\in(0,1]$, provided the teacher variance distribution over $\mathcal{X}$ has positive mass in $(1-c,1)$.
\end{theorem}

\begin{assumption}[Bounded teacher noise]\label{assum:noise}
For each unlabeled sample $x_i$, write $\mathbf{q}_i^{\mathcal T}=\mathbf{q}_i^{\star}+\boldsymbol{\varepsilon}_i$, where $\mathbf{q}_i^{\star}$ is the Bayes-optimal non-target distribution and $\boldsymbol{\varepsilon}_i$ is zero-mean noise satisfying $\Vert\boldsymbol{\varepsilon}_i\Vert_2\le C\sigma_i^{\mathcal T}$ for some constant $C>0$. Moreover, $\mathbb{E}[\Vert\boldsymbol{\varepsilon}_i\Vert_2^2\mid\sigma_i^{\mathcal T}]$ is non-decreasing in $\sigma_i^{\mathcal T}$, so the teacher variance serves as a calibrated reliability proxy~\cite{rizve2021defense,sun2024heteroscedastic}.
\end{assumption}

\begin{proof}[Proof of Theorem~\ref{thm:main}]
By the $G$-Lipschitz condition of the KL gradient on the bounded simplex used by DKD,
$\Vert\omega_i\nabla_\theta\mathrm{KL}(\mathbf{q}_i^{\mathcal T}\Vert\mathbf{q}_i^{\mathcal S})
-\omega_i\nabla_\theta\mathrm{KL}(\mathbf{q}_i^{\star}\Vert\mathbf{q}_i^{\mathcal S})\Vert_2
\le \omega_i G \Vert\boldsymbol{\varepsilon}_i\Vert_2
\le \omega_i G C \sigma_i^{\mathcal T}$.
Squaring and taking expectation yields Eq.~\eqref{eq:noise-bound}.
Since $\omega_i=1-\sigma_i^{\mathcal T}<1$ whenever $\sigma_i^{\mathcal T}>0$, the bound is strictly tighter than the unweighted case.
\end{proof}

\begin{proof}[Proof of Theorem~\ref{thm:main1}]
Define per-sample loss:
\begin{align}
h_i(\theta)=\omega_i\mathrm{KL}(\mathbf{q}_i^{\mathcal T}\Vert\mathbf{q}_i^{\mathcal S}).
\end{align}

Because KL is bounded on the compact simplex, $|h_i|\le\omega_i B$ for some $B>0$. Applying McDiarmid's inequality with bounded-difference coefficients $\omega_i B/|\mathcal X|$ and the standard Rademacher symmetrization gives Eq.~\eqref{eq:risk-bound}. For any fixed constant weight $c$, on the set $\{\sigma_i^{\mathcal T}>1-c\}$, we have $(1-\sigma_i^{\mathcal T})^2<c^2$, so $\sum_i\omega_i^2(\sigma_i^{\mathcal T})^2<\sum_i c^2(\sigma_i^{\mathcal T})^2$, making the adaptive noise term strictly smaller.
\end{proof}

In short, $\omega_i$ multiplies the noise budget by a factor that vanishes as $\sigma_i^{\mathcal T}\to 1$, so the weight can largely remove more noise than static. Thus, the weighting design for non-target samples mitigates the negative impact of uncertain samples while placing greater emphasis on reliable teacher guidance. The empirical study in Section~\ref{sec:ablation-weight} further validates this design.


\subsection{Training and Inference}
\noindent\textbf{Training.} During training, DKD jointly trains the
teacher and student with
\begin{equation}
\mathcal{L}_{DKD}=\mathcal{L}_{\mathcal T}+\mathcal{L}_{\mathcal S}+\mathcal{L}_{DDA}.
\end{equation}

\noindent\textbf{Inference.} %
During inference, we deploy \emph{only the student model}, i.e.
\begin{equation}
\hat y_i=\sum_{i=1}^{L}p_i^{\mathcal S}\,\mathbf{b}_i.
\end{equation}
The teacher branch is purely a training-time device for producing
cleaner distributional supervision and is discarded at deployment.
As a consequence, DKD's inference-time footprint is identical to
that of a single-model baseline of the same backbone.
Section~\ref{sec:efficiency} reports per-sample latency numbers
showing that DKD is \emph{faster} than RankUp and GCLSS on
UTKFace, and on par with them on BVCC.

\section{Experiments}\label{sec:experiments}
\subsection{Datasets}
We conduct experiments on four real-world
regression benchmarks spanning speech, text, computer vision, and clinical data analysis. Table~\ref{tab:datasets} summarizes the key
statistics, and we give a brief description of each below.

\begin{itemize}
    \item \textbf{BVCC}~\cite{cooper2021voices} is an audio quality
assessment benchmark whose labels are averaged scores from multiple
human listeners on a 1-to-5 Likert scale, with 4{,}974 training,
1{,}066 validation, and 1{,}066 test samples. Following the protocol
of RankUp~\cite{huang2024rankup}, we use only the training and
validation splits for evaluation.

 \item \textbf{Yelp}~\cite{asghar2016yelp} is a textual opinion-mining
task predicting customer ratings. We use the preprocessed split from
USB~\cite{wang2022usb} (250k train, 25k validation, 10k test) and
report results on the validation set, consistent with RankUp.

 \item \textbf{UTKFace}~\cite{zhang2017age} addresses image-based age
estimation with labels in $[1,116]$ and a long-tailed distribution.
RankUp's split gives 18{,}964 training and 4{,}741 test samples on
the aligned--cropped images.

 \item \textbf{MIMIC}~\cite{johnson2023mimic} is a multivariate
time-series clinical task predicting SOFA severity scores in
$[0,24]$, with 3{,}623{,}503 training and 55{,}859 test samples over
multivariate ICU features.
Together these four datasets cover three modalities (image, audio, text) plus multivariate medical time series, two label scales (small-range 1--5 vs.\ large-range
1--116), and both balanced and long-tailed label distributions,
which lets us stress-test the generality of DKD.
\end{itemize}
\begin{table}[!t]
\centering
\small
\caption{The statistical information of the four datasets.}
\label{tab:datasets}
\setlength{\tabcolsep}{0.60mm}
\begin{tabularx}{\linewidth}{c*{4}{>{\centering\arraybackslash}X}}
\toprule
Model           & BVCC & Yelp  & UTKFace & MIMIC \\ \midrule
\# Train Set    & 4{,}974 & 250{,}000 & 18{,}964 & 3{,}623{,}503 \\
\# Test Set     & 1{,}066 & 25{,}000  & 4{,}741  & 55{,}859 \\
Domain          & Audio   & Text      & Image    & Medical  \\
\bottomrule
\end{tabularx}
\end{table}

\subsection{Competing Methods}\label{sec:experiments-competing} 

We compared our DKD model with several SOTA models that fall into three groups.

(1) \textbf{Direct regression (DR)}: A vanilla supervised baseline that trains the same backbone as the semi-supervised methods directly on the labeled subset by minimizing MAE loss, without incorporating the unlabeled data.

(2) \textbf{Pseudo-label and
consistency-regularization methods}:\noindent \textbf{\( \pi \)-Model} \cite{laine2016temporal} achieves self-ensembling by aggregating the outputs of a single network at different training stages. Consensus predictions for unknown labels are formed through diverse regularization and augmentation strategies. Perturbations are injected into both branches, and consistency between the two perturbed outputs is enforced, ensuring a more balanced and symmetric comparison.
\noindent \textbf{Mean Teacher} \cite{tarvainen2017mean} aims to train a model via consistency between original samples and their perturbed counterparts, while using an exponential-moving-average (EMA) version of the model—termed the teacher—to supply more reliable guidance to the current student during training.
\noindent \textbf{MixMatch} \cite{berthelot2019mixmatch} consolidates the dominant approaches in semi-supervised learning by generating low-entropy label guesses for augmented unlabeled examples and subsequently blending labeled and unlabeled data via MixUp.
\noindent \textbf{UCVME} \cite{dai2023ucvme} improves training by producing high-quality uncertainty estimates for pseudo-labels via heteroscedastic regression. By explicitly modeling label uncertainty, the approach assigns greater importance to more reliable pseudo-labels. Furthermore, a novel variational model-ensembling scheme is introduced to reduce predictive noise and generate stronger pseudo-labels.


(3) \textbf{Contrastive-learning-based methods}: 
\noindent \textbf{CLSS} \cite{dai2023semi} extends contrastive regression to the semi-supervised domain. It exploits a spectral ordering algorithm to extract ordinal rankings from the feature-similarity matrix of unlabeled data and then leverages these rankings as supervisory signals. The additional ranking loss enhances robustness, marking the first exploration of contrastive learning for regression tasks.
\noindent \textbf{GCLSS}~\cite{wang2025contrastive} extends CLSS to allow both labeled
and unlabeled data to participate in the similarity matrix and uses
the recovered rankings as pseudo-labels;
\noindent \textbf{RankUp} \cite{huang2024rankup} reframes the original regression task as a ranking problem and trains it jointly with the regression objective. By introducing an auxiliary ranking classifier whose outputs can be handled by any off-the-shelf semi-supervised classification method, RankUp seamlessly integrates existing classification techniques into regression-oriented semi-supervised learning.

DKD differs from all three groups: instead of self-training, ordinal
mining or auxiliary ranking, it transfers calibrated continuous and
distributional knowledge from a label-only-trained teacher.

\noindent\textbf{Evaluation Metric.} Following
\cite{huang2024rankup,dai2023semi}, we adopt MAE, the coefficient of
determination $R^2$, and the Spearman rank correlation (SRCC):
\[
\mathrm{MAE}=\tfrac{1}{N}\!\sum_{i=1}^{N}|\hat y_i-y_i|,
\quad
R^2=1-\tfrac{\sum_i(\hat y_i-y_i)^2}{\sum_i(\bar y-y_i)^2},
\quad
\mathrm{SRCC}=\tfrac{\mathrm{cov}(r(\hat y),r(y))}{\sigma_{r(\hat y)}\sigma_{r(y)}},
\]
where $r(\cdot)$ denotes ranks and $\bar y$ is the label mean. MAE
measures absolute error, $R^2\in(0,1]$ measures explained
variance, and $\mathrm{SRCC}\in[-1,1]$ captures rank consistency.

\noindent\textbf{Implementation Details.} DKD is implemented in
Python 3.10 with PyTorch 2.1.2 on an NVIDIA GeForce RTX 4060 (8\,GB).
Backbones follow~\cite{huang2024rankup}: Whisper-Base for BVCC,
Bert-Small for Yelp, Wide-ResNet-28-2 for UTKFace, and Bert for
MIMIC. We set $\beta=10$ and $L=200$ for BVCC, Yelp, and UTKFace, and
$L=100$ for MIMIC. To make the experiments fully reproducible we
report the full training configuration of every backbone in
Table~\ref{tab:hyperparameters1}, and the per-dataset weak/strong
augmentation schemes in Table~\ref{tab:augmentation}. For the three
natural-modality datasets (BVCC, Yelp, UTKFace) we reuse the
augmentation policies of the universal semi-supervised learning
benchmark USB~\cite{wang2022usb}, so that DKD and all competing
methods share identical augmentation. For MIMIC, whose inputs are
multivariate time series, we define weak augmentation as random
masking of $5\%$ observed entries and strong augmentation as $m$
rounds of random masking combined with additive Gaussian noise
($\mathcal N(0,0.02)$) per variate, following the treatment in
\cite{huang2024rankup}.

\begin{table}[t]
\centering
\footnotesize
\caption{Default hyperparameters of the four base models. 
}
\label{tab:hyperparameters1}
\setlength{\tabcolsep}{0.9mm}
\renewcommand{\arraystretch}{1.10}
\resizebox{0.45\textwidth}{!}{%
\begin{tabular}{lcccc}
\toprule
Models & \makecell{Wide\\ResNet-28-2} & \makecell{Whisper\\Base} & \makecell{Bert\\Small} & Bert \\
\midrule
Training Iterations   & 262{,}144 & 102{,}400 & 102{,}400 & 1{,}024{,}000 \\
Evaluation Iterations & 1{,}024   & 1{,}024   & 1{,}024   & 1{,}024       \\
Batch Size            & 32        & 8         & 8         & 8 \\
Optimizer             & SGD       & AdamW     & AdamW     & AdamW \\
Momentum              & 0.9       & --        & --        & -- \\
Criterion             & MAE       & MAE       & MAE       & MAE \\
Weight Decay          & 1e-03     & 2e-05     & 5e-04     & 2e-5 \\
Layer Decay           & 1.0       & 0.75      & 0.75      & 0.75 \\
Learning Rate         & 1e-02     & 2e-06     & 1e-05     & 2e-06 \\
EMA Weight            & 0.999     & --        & --        & 0.999 \\
Pretrained            & False     & True      & True      & False \\
Image Resize          & 40x40     & --        & --        & -- \\
Max Length Seconds    & --        & 6.0       & --        & -- \\
Sample Rate           & --        & 16{,}000  & --        & -- \\
Max Length            & --        & --        & 512       & -- \\
\bottomrule
\end{tabular}
}
\end{table}

\begin{table}[t]
\centering
\small
\caption{Weak and strong augmentation policies for each dataset.
}
\label{tab:augmentation}
\setlength{\tabcolsep}{3pt}
\resizebox{0.45\textwidth}{!}{%
\begin{tabular}{@{}ll@{}}
\toprule
\textbf{Dataset} & \textbf{Augmentation} \\
\midrule
UTKFace &
  \textbf{Weak}: Random Crop, Random Horizontal Flip \\
  & \textbf{Strong}: RandAugment~\cite{cubuk2020randaugment} \\[2pt]
BVCC &
  \textbf{Weak}: Random Sub-sample \\
  & \textbf{Strong}: Random Sub-sample, Mask, Trim, Padding \\[2pt]
Yelp &
  \textbf{Weak}: None \\
  & \textbf{Strong}: Back-Translation~\cite{xie2020unsupervised} \\[2pt]
MIMIC &
  \textbf{Weak}: Random Masking (5\%) \\
  & \textbf{Strong}: $m$ Masking rounds $+$ Gaussian noise $\mathcal N(0,0.02)$ \\
\bottomrule
\end{tabular}
}
\end{table}

\subsection{Main Comparison Results}\label{sec:main-results}

\noindent\textbf{Protocol.} Unless otherwise stated we follow the
standard SSR protocol of RankUp~\cite{huang2024rankup} and
USB~\cite{wang2022usb}: 250 labeled samples are sampled per dataset
with fixed random seeds and the remaining training set is used as
unlabeled data; every method is trained for the iteration budget
reported in Table~\ref{tab:hyperparameters1}; results are averaged
over the last 16 evaluation iterations to stabilize the rankings.
We use the same labeled/unlabeled partition for DKD and all
competing methods to ensure a fair comparison.

Table~\ref{tab:main} summarizes the performance of all methods across
the four datasets. All semi-supervised methods are trained with 250
labeled samples. DKD significantly outperforms the baselines on all
datasets. Specifically, DKD outperforms the second-best model, GCLSS,
by 4.97\% in MAE, 11.8\% in $R^2$, and 8.04\% in SRCC on average.
GCLSS degrades significantly on MIMIC, because its heavy reliance on
the feature similarity matrix is sensitive to the noise inherent in
real-world ICU data. In contrast, DKD achieves state-of-the-art
performance by fully leveraging unlabeled samples through DDA.

\begin{table*}[t]
\centering
\small
\caption{Main comparison on four datasets. Best is bold,
second-best underlined.}
\label{tab:main}
\setlength{\tabcolsep}{1.10mm}
\renewcommand{\arraystretch}{1.15}
\resizebox{\textwidth}{!}{%
\begin{tabular}{ccccccccccccc}
\toprule
\multirow{2}{*}{Model}
& \multicolumn{3}{c}{BVCC}
& \multicolumn{3}{c}{Yelp}
& \multicolumn{3}{c}{UTKFace}
& \multicolumn{3}{c}{MIMIC}\\
\cmidrule(lr){2-4}\cmidrule(lr){5-7}\cmidrule(lr){8-10}\cmidrule(lr){11-13}
& MAE $\downarrow$ & $R^2$ $\uparrow$ & SRCC $\uparrow$
& MAE $\downarrow$ & $R^2$ $\uparrow$ & SRCC $\uparrow$
& MAE $\downarrow$ & $R^2$ $\uparrow$ & SRCC $\uparrow$
& MAE $\downarrow$ & $R^2$ $\uparrow$ & SRCC $\uparrow$ \\
\midrule
DR                 & 0.533 & 0.490 & 0.741 & 0.723 & 0.566 & 0.769 & 9.420 & 0.540 & 0.712 & 2.726 & 0.239 & 0.510 \\
$\pi$-Model (2017) & 0.534 & 0.489 & 0.740 & 0.730 & 0.565 & 0.769 & 9.450 & 0.534 & 0.706 & 2.725 & 0.240 & 0.506 \\
Mean Teacher (2017)& 0.532 & 0.492 & 0.742 & 0.730 & 0.565 & 0.769 & 8.850 & 0.586 & 0.745 & 2.724 & 0.240 & 0.507 \\
MixMatch (2019)    & 0.597 & 0.353 & 0.626 & 0.886 & 0.381 & 0.660 & 7.950 & 0.692 & 0.832 & 2.707 & 0.252 & 0.514 \\
CLSS (2023)        & 0.499 & 0.534 & 0.748 & 0.721 & 0.543 & 0.748 & 9.100 & 0.586 & 0.737 & 2.714 & 0.251 & 0.510 \\
UCVME (2023)       & 0.498 & 0.553 & 0.774 & 0.775 & 0.540 & 0.763 & 8.630 & 0.626 & 0.767 & \underline{2.656} & \underline{0.285} & \underline{0.539} \\
RankUp (2024)      & 0.470 & 0.588 & 0.776 & 0.661 & 0.645 & \underline{0.829} & 7.060 & 0.751 & 0.835 & 2.720 & 0.244 & 0.509 \\
GCLSS (2026)       & \underline{0.455} & \underline{0.613} & \underline{0.786}
                   & \underline{0.592} & \underline{0.669} & 0.821
                   & \underline{7.018} & \underline{0.743} & \underline{0.829}
                   & 2.934 & 0.135 & 0.374 \\
DKD (Ours)         & \textbf{0.422} & \textbf{0.669} & \textbf{0.818}
                   & \textbf{0.549} & \textbf{0.694} & \textbf{0.839}
                   & \textbf{6.852} & \textbf{0.761} & \textbf{0.838}
                   & \textbf{2.629} & \textbf{0.291} & \textbf{0.541} \\
\bottomrule
\end{tabular}}
\end{table*}

Consistency regularization-based methods ($\pi$-Model, Mean Teacher,
MixMatch) consistently show inferior performance, mostly matching DR.
Hybrid augmentation struggles to generate high-quality numeric
pseudo-labels for regression, particularly with extremely limited
labels. DKD's decoupled distillation both enhances pseudo-label
quality and more effectively utilizes target and non-target
distributions. As for the co-trained UCVME, which maximizes
variational mutual information, it is suboptimal on MIMIC and
requires a more complex Bayesian neural network. DKD's dual-stream
distillation consistently surpasses UCVME and CLSS.

\noindent\textbf{Cross-modality analysis.}
This comparison asks whether DKD helps equally across modalities or
whether its gain depends on the data regime. Table~\ref{tab:main}
shows that DKD improves over the strongest baseline on all four
benchmarks, but the margin is not uniform. On BVCC, the MAE drops from
$0.455$ to $0.422$. On UTKFace, the gain is smaller
in relative terms, from $7.018$ to $6.852$, though it is
obtained on the widest label range in the paper. On Yelp, DKD reduces
MAE from $0.592$ to $0.549$. The largest absolute gap appears
on MIMIC, where DKD reaches $2.629$ while the strongest competing
method remains at $2.656$, and GCLSS drops to $2.934$.

Taken together, these results suggest that DKD is not tied to a single
modality. The gain is moderate on the easier regimes (BVCC), remains
stable on the long-tailed vision task (UTKFace), and becomes more
visible on the harder clinical setting (MIMIC), where pseudo-label
quality is likely to matter more. 

\subsection{Ablation Study}
\begin{table*}[t]
\centering
\caption{Ablation study on the proposed components. Best in bold,
second-best underlined.}
\label{tab:ablation}
\setlength{\tabcolsep}{1.40mm}
\renewcommand{\arraystretch}{1.15}
\resizebox{\textwidth}{!}{%
\begin{tabular}{ccccccccccccc}
\toprule
\multirow{2}{*}{Model}
& \multicolumn{3}{c}{BVCC}
& \multicolumn{3}{c}{Yelp}
& \multicolumn{3}{c}{UTKFace}
& \multicolumn{3}{c}{MIMIC}\\
\cmidrule(lr){2-4}\cmidrule(lr){5-7}\cmidrule(lr){8-10}\cmidrule(lr){11-13}
& MAE $\downarrow$ & $R^2$ $\uparrow$ & SRCC $\uparrow$
& MAE $\downarrow$ & $R^2$ $\uparrow$ & SRCC $\uparrow$
& MAE $\downarrow$ & $R^2$ $\uparrow$ & SRCC $\uparrow$
& MAE $\downarrow$ & $R^2$ $\uparrow$ & SRCC $\uparrow$ \\
\midrule
S-LDL      & \underline{0.441} & \underline{0.640} & \underline{0.795}
           & 0.717 & 0.514 & 0.728
           & 9.570 & 0.505 & 0.687
           & 2.748 & 0.227 & 0.481 \\
DKD-Logits & 0.506 & 0.529 & 0.737
           & \underline{0.576} & \underline{0.655} & \underline{0.819}
           & 7.981 & 0.648 & 0.799
           & 2.707 & 0.248 & 0.520 \\
DKD-KL     & 0.500 & 0.547 & 0.747
           & 0.610 & 0.617 & 0.803
           & \underline{7.502} & \underline{0.693} & \underline{0.808}
           & \underline{2.674} & \underline{0.264} & \underline{0.531} \\
DKD (Ours) & \textbf{0.422} & \textbf{0.669} & \textbf{0.818}
           & \textbf{0.549} & \textbf{0.694} & \textbf{0.839}
           & \textbf{6.852} & \textbf{0.761} & \textbf{0.838}
           & \textbf{2.629} & \textbf{0.291} & \textbf{0.541} \\
\bottomrule
\end{tabular}}
\end{table*}
\noindent\textbf{Ablation on the Effort of Components.}
We evaluate the contribution of each component in DKD by designing
the following three variants.

(1) \textbf{S-LDL}, which removes the distillation
framework and keeps only one branch trained on labeled samples;

(2)\textbf{DKD-Logits}, which removes DDA and keeps only label alignment; 

(3) \textbf{DKD-KL}, which replaces DDA with standard KL divergence.

The results are shown in Table \ref{tab:ablation}. From the results, the DKD can consistently outperform all the variants and yields the best performance on all the datasets, demonstrating its effectiveness.  
Both DKD-KL and DKD-Logits underperform the full DKD model but outperform S-LDL on the Yelp, UTKFace, and MIMIC datasets. This demonstrates that distillation effectively captures generalized patterns across text, image, and tabular medical data modalities.
On the BCVV dataset, S-LDL achieves the second-best performance, though still behind DKD. This suggests that while S-LDL, leveraging only label distribution learning, can effectively learn patterns in audio data, the absence of distillation and DDA limits its ability, leading to suboptimal results. 
The full DKD consistently achieves the best performance, confirming the
effectiveness of both the distillation framework and DDA.

\noindent\textbf{Ablation on the Variance-aware Weight.}
\label{sec:ablation-weight}
This ablation tests the specific role of the sample-wise weight in Eq.~\eqref{eq:weight}.  
The question is not simply whether a smaller non-target KL helps, but whether the gain of DDA comes from \emph{adapting} the weight to each sample's uncertainty.  
To answer this, we replace $\omega_i=1-\sigma_i^{\mathcal T}$ with four alternatives and keep all other DKD components fixed: unit weight $\omega= 1$, three constants $\omega= c$ with $c\in\{0.2,0.5,0.8\}$.

Table~\ref{tab:var-weight} shows that the adaptive rule gives the lowest MAE on all four datasets. 
Compared with the strongest non-adaptive alternative on each dataset, it reduces MAE from $0.433$ to $0.422$ on BVCC, from $0.623$ to $0.549$ on Yelp, from $7.234$ to $6.852$ on UTKFace, and from $2.872$ to $2.629$ on MIMIC.  
A second pattern is that no single constant works best across datasets: the closest non-adaptive competitor is the $\omega= 0.2$ on BVCC, Yelp and MIMIC, and $\omega= 1$ on UTKFace.  
This instability suggests that the benefit does not come from uniformly shrinking the non-target KL by a fixed amount; it comes from assigning less trust to the teacher precisely on high-variance samples.

The gap is largest on UTKFace and MIMIC, where the label range is broader and teacher uncertainty is likely to vary more across samples.  
This empirical trend is consistent with Theorem~\ref{thm:main}: when the variance spread is larger, a sample-wise rule has more room to improve over any fixed coefficient.

\begin{table}[t]
\centering
\small
\caption{Non-target weighting schemes on all four datasets (MAE,
lower is better). Variance-aware Weight vs. static weight. Best in bold.}
\label{tab:var-weight}
\setlength{\tabcolsep}{1.4mm}
\begin{tabular}{lcccc}
\toprule
Weighting scheme for non-target KL
  & BVCC & Yelp & UTKFace & MIMIC\\
\midrule
Unit weight $\omega= 1$           & 0.473 & 0.628 & 7.234 & 2.933 \\
Static $\omega= 0.2$              & 0.443 & 0.623 & 7.517 & 2.872 \\
Static $\omega= 0.5$              & 0.446 & 0.631 & 7.274 & 2.915 \\
Static $\omega= 0.8$              & 0.457 & 0.629 & 7.300 & 2.929 \\
Adaptive $\omega_i\!=\!1\!-\!\sigma_i^{\mathcal T}$~(\textbf{ours})
  & \textbf{0.422} & \textbf{0.549} & \textbf{6.852} & \textbf{2.629} \\
\bottomrule
\end{tabular}
\end{table}

\begin{figure}[t]
  \centering
  \includegraphics[width=1.0\linewidth]{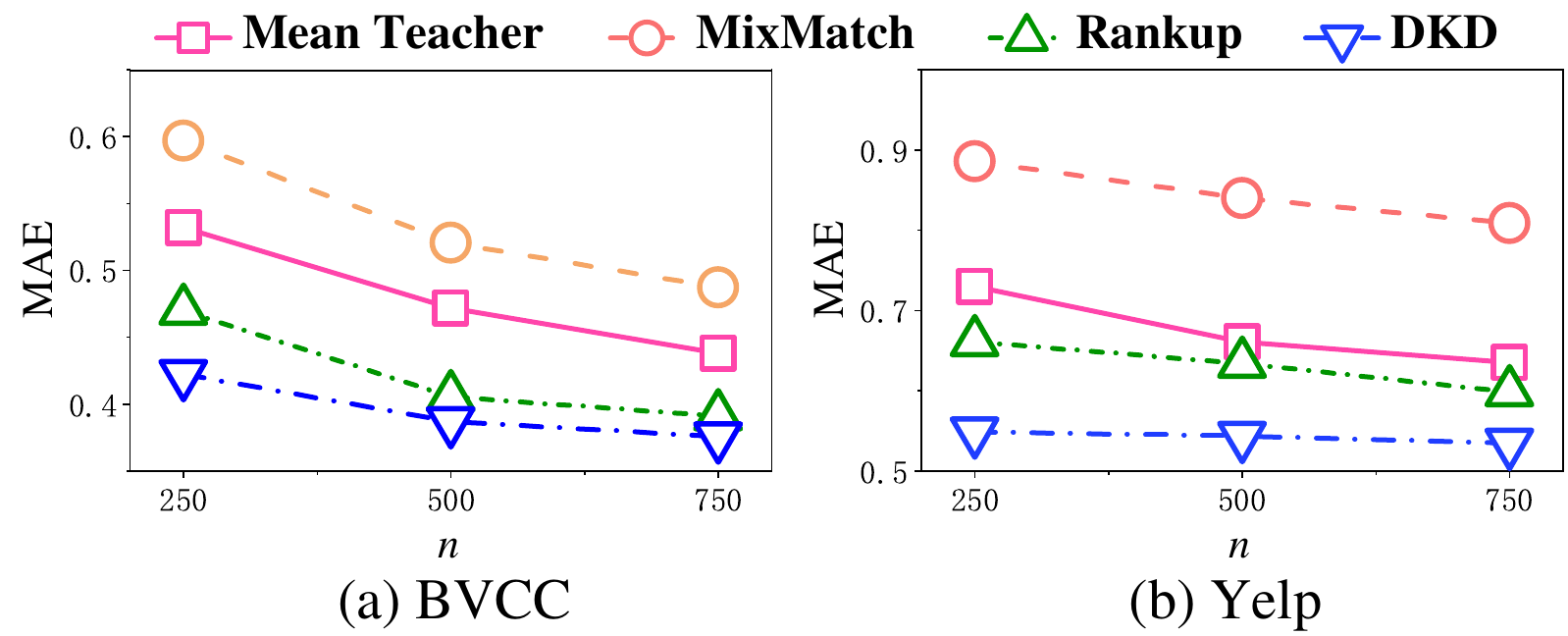}
  \caption{MAE w.r.t. the labeled sample size $n$.}
  \label{fig:labeled}
\end{figure}

\begin{figure}[t]
  \centering
  \includegraphics[width=1.0\linewidth]{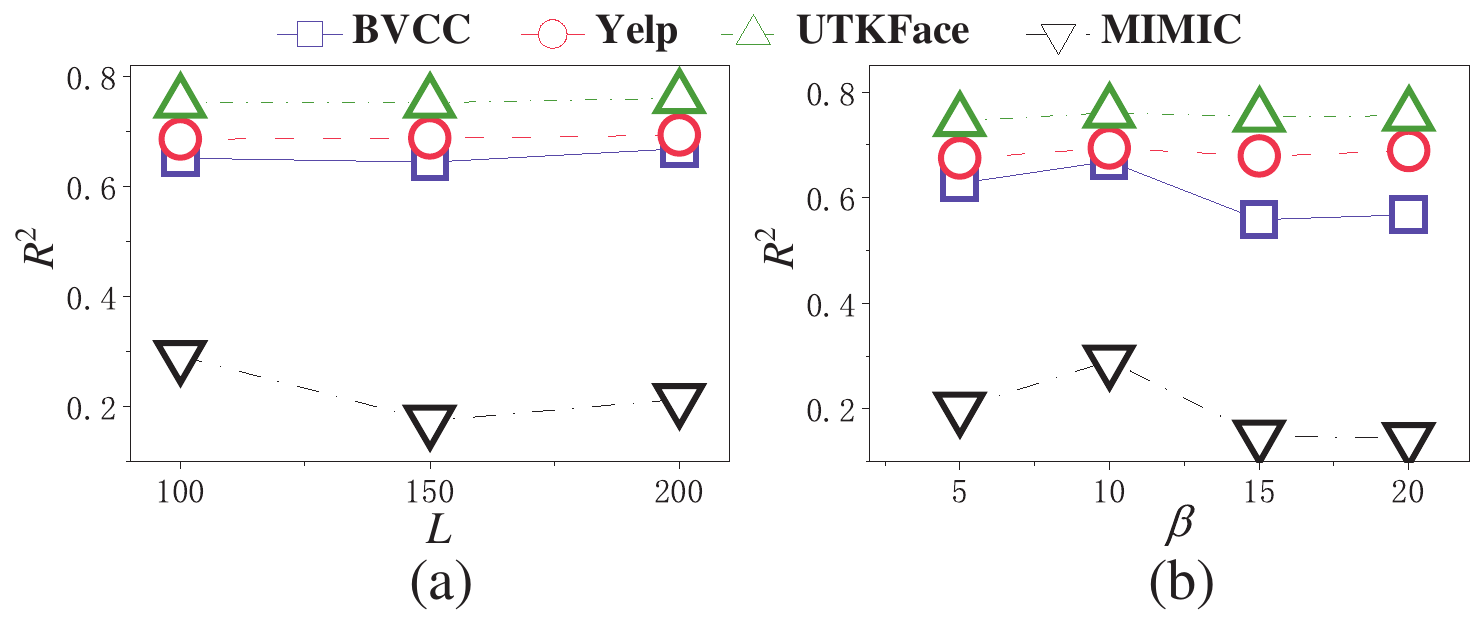}
  \caption{Effectiveness of the hyperparameters $L$ and $\beta$.}
  \label{fig:hyper}
\end{figure}

\subsection{Performance w.r.t. Labeled Sample Size}
Figure~\ref{fig:labeled} reports MAE across different labeled
sizes. As $n$ grows, DKD generally improves on all four datasets, and
consistently achieves the best performance at every labeling ratio,
confirming its robustness.

\subsection{Sensitivity Analysis}
\begin{figure}[t]
  \centering
  \begin{subfigure}{1\linewidth}
    \includegraphics[width=\linewidth]{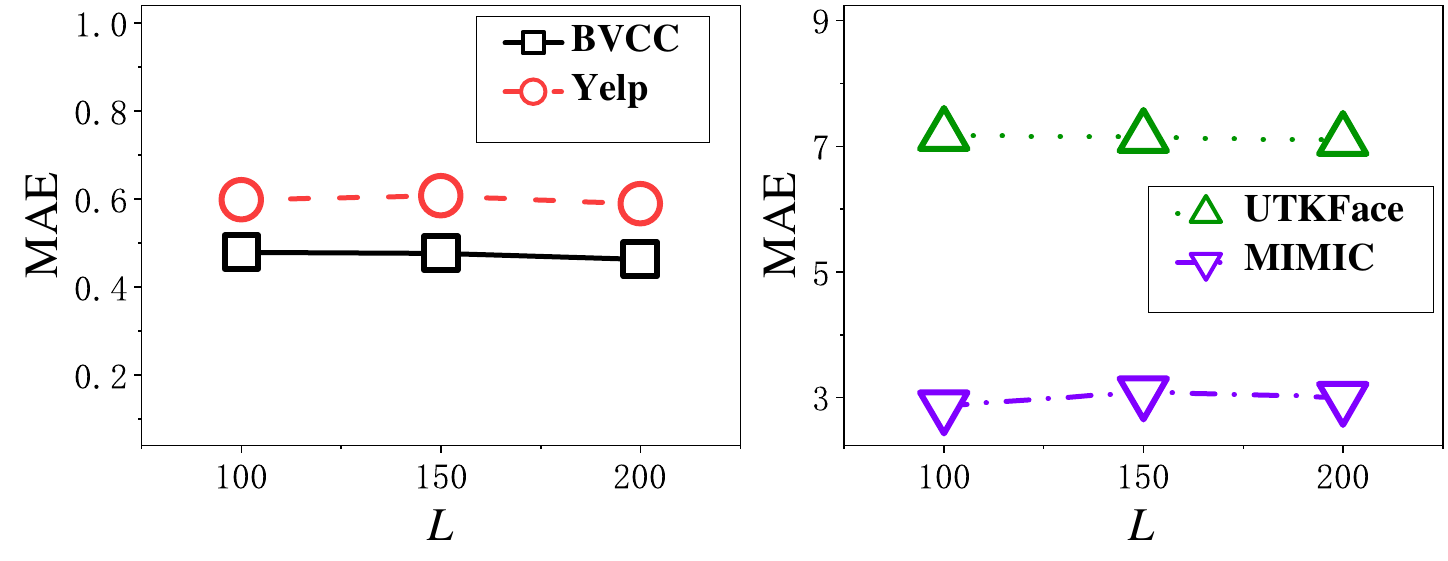}
    \caption{The MAE w.r.t of buckets size $L$.}
    \label{fig:mae_L_main}
  \end{subfigure}
  \hfill
  \begin{subfigure}{1\linewidth}
    \includegraphics[width=\linewidth]{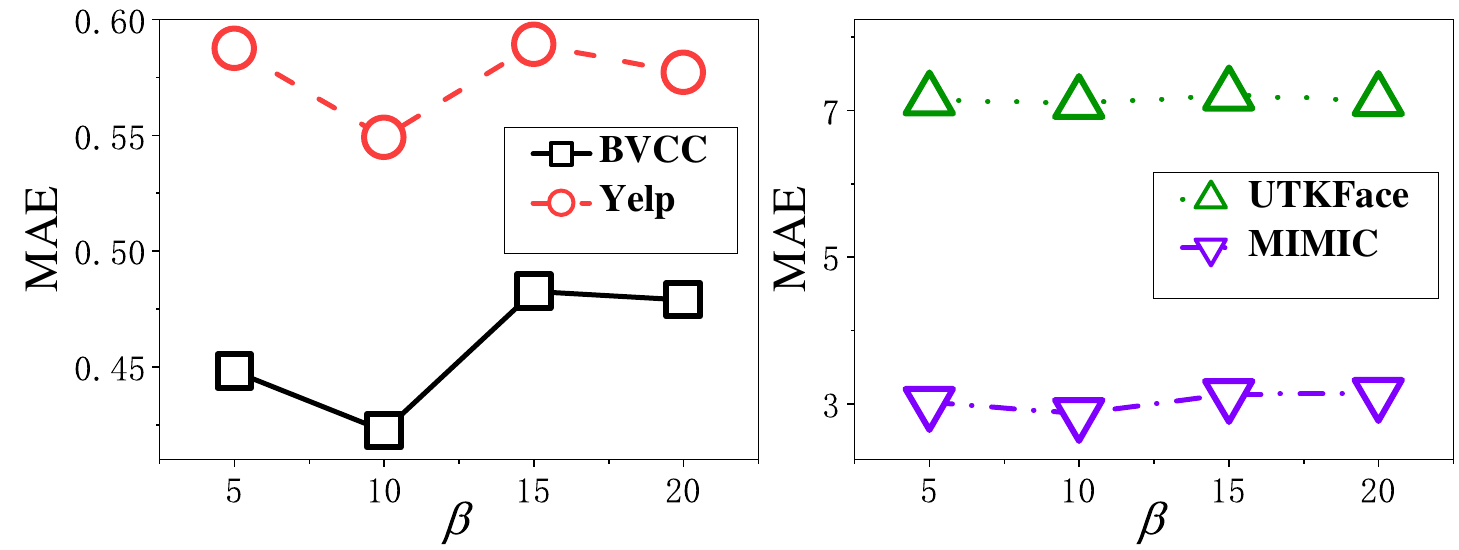}
    \caption{The MAE w.r.t of $\beta$.}
    \label{fig:mae_beta_main}
  \end{subfigure}
  \caption{
   The effectiveness of the hyperparameters $\beta$.
  }
  \label{fig:mae-main}
\end{figure}

We examine the sensitivity of DKD w.r.t. $L$ and $\beta$. The $R^2$
results are in Figure~\ref{fig:hyper}, and the MAE counterparts in
Figure~\ref{fig:mae-main}.

\noindent\textbf{Effect of $L$.} DKD performs stably under different
$L$ on text, image, and audio, with a slight improvement as $L$
grows, because a larger $L$ captures the label distribution more
accurately. On MIMIC, $L=100$ is best, primarily because SOFA scores
are concentrated within a narrow interval. MAE is
even more stable than $R^2$ under varying $L$, which indicates that
DKD is robust to the discretization granularity.

\noindent\textbf{Effect of $\beta$ in $\mathcal{L}_{DDA}$.} Across all
datasets, $\beta=10$ yields optimal performance. A smaller $\beta$
over-emphasizes difficult samples, while a larger $\beta$ over-biases
the student toward the non-target signal; $\beta=10$ strikes the right
balance.


\subsection{Distribution Visualization}

\begin{figure}[t]
  \centering
  \begin{subfigure}{0.48\linewidth}
    \includegraphics[width=\linewidth]{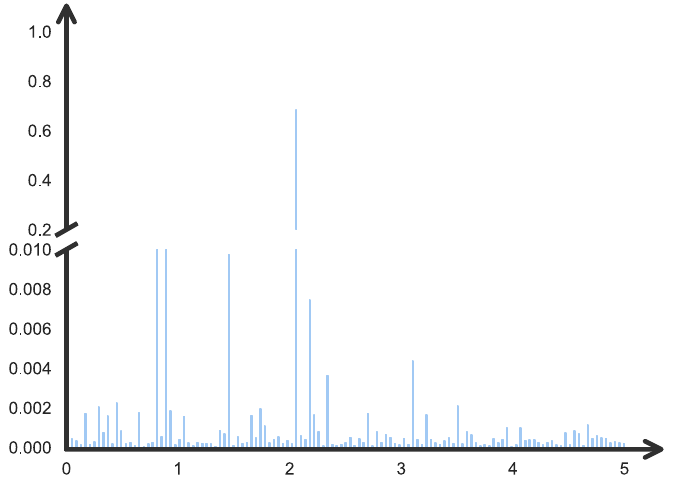}
    \caption{Ground truth $y=4$.}
    \label{fig:vis-a}
  \end{subfigure}
  \hfill
  \begin{subfigure}{0.48\linewidth}
    \includegraphics[width=\linewidth]{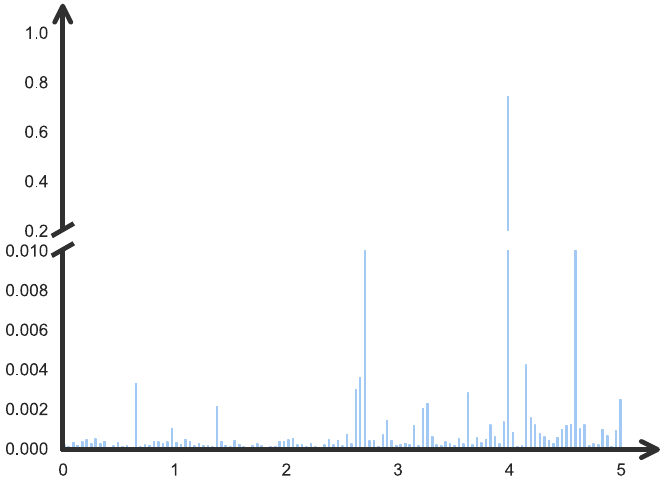}
    \caption{Ground truth $y=2$.}
    \label{fig:vis-b}
  \end{subfigure}
  \caption{Visualization of DKD's predictions on Yelp for two samples with ground-truth labels $y=2$ and $y=4$.}
  \label{fig:vis}
\end{figure}
\begin{figure*}[t]
  \centering
  \includegraphics[width=\linewidth]{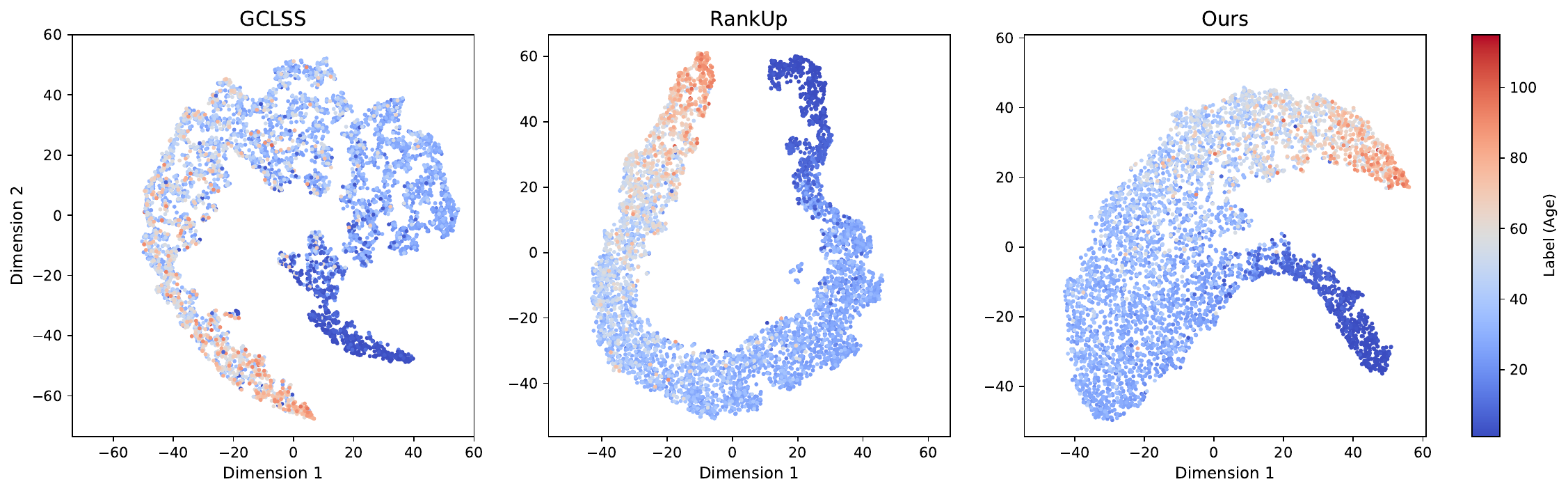}
  \caption{t-SNE visualization of feature embeddings on UTKFace,
  coloured by ground-truth age.}
  \label{fig:tsne-utkface}
\end{figure*}
Figure~\ref{fig:vis} shows two prediction examples on Yelp. The model produces a smooth distribution that concentrates around the ground-truth bin while still spreading mass over neighbouring ordinal bins, demonstrating that the non-target stream preserves the ordinal structure of the continuous label space. In
Figure~\ref{fig:vis}(a), for ground-truth value $4$, the predicted
distribution concentrates around $4$ but leaves residual mass on the
adjacent values $3$ and $5$; an analogous pattern is visible in
Figure~\ref{fig:vis}(b). This empirically verifies the claim made in
Section~\ref{sec:var-weight}: although DDA decouples the target and
non-target distributions, the non-target branch keeps the ordinal
structure of the bin space, and the final regression output
$\hat y=\sum_i p_i\mathbf{b}_i$ smoothly aggregates this structured
mass over neighbouring bins.

\subsection{Feature-space Geometry: Smoothness and Low-density
on UTKFace}\label{sec:tsne}

The decoupled distribution alignment in DKD imposes \emph{two}
distinct supervisory pressures on the student: the target branch
pulls together samples whose ground-truth (or pseudo) labels fall
in the same bin, and the non-target branch transfers an ordered
soft profile that pushes samples with dissimilar labels into
different positions of the bin space. We adopt the generalized interpretation of the smoothness and low-density assumptions introduced in \cite{huang2024rankup}. Specifically, the smoothness assumption encourages features with similar labels to cluster, whereas the low-density assumption enforces separation among features with disparate labels. Consequently, we expect the penultimate features of DKD to exhibit both dynamics within a regression manifold.

Figure~\ref{fig:tsne-utkface} shows t-SNE projections of the
student's penultimate features on UTKFace, coloured by the
continuous age label. GCLSS produces a fragmented embedding in
which young (blue) and older (red) ages are scattered into multiple
disjoint clusters; the manifold lacks a clear age-ordered axis.
RankUp recovers a more globally ordered manifold thanks to its
pairwise-ranking auxiliary, but mid-age samples remain mixed with
the high-age tail and the cluster boundary on the right are loose.
DKD yields the most label-coherent geometry of the three: features
of similar age are tightly grouped (smoothness), and the colour
gradient transitions monotonically from blue (low age) to red (high
age) along a single arc with a clearly separated dark-blue tail
(low-density). This is consistent with our theoretical analysis in
Section~\ref{sec:var-weight}: the variance-aware non-target
weighting strips noisy mass from teacher predictions on uncertain
unlabeled samples, so the student receives cleaner ordinal
supervision and learns a representation that simultaneously honours
the smoothness and low-density assumptions in their
regression-adapted form.

\subsection{Efficiency Analysis}\label{sec:efficiency}
\noindent\textbf{Time complexity.} Given that both models share the
same backbone output, we exclude the backbone cost. (1) The label
distribution learning module is a fully connected layer plus an
expectation, with time complexity $\mathcal O(LM+2L)$, where $M$ is
the backbone output feature dimension. (2) DKD contains DDA and
continuous-valued distillation with complexity $\mathcal O(L)$ and
$\mathcal O(1)$ per sample respectively. 
In particular, increasing $L$ to achieve higher regression precision does not introduce any nonlinear computational overhead. For instance, raising $L$ from $100$ to $200$ strictly doubles the floating-point operations (FLOPs) within the LDL head, a marginal increase that is negligible compared to the computational demands of the backbone network.

\begin{table}[t]
\centering
\small
\caption{Per-sample inference time (ms) on the four datasets. Batch size $1$, averaged over $1{,}000$ runs after $100$ warm-ups.}
\label{tab:inference}
\setlength{\tabcolsep}{1.4mm}
\resizebox{0.45\textwidth}{!}{%
\begin{tabular}{lcccc}
\toprule
Model         & UTKFace & BVCC & Yelp & MIMIC \\
\midrule
UCVME (2023)  & 46.52 & 66.25 & 44.67  & 25.05 \\
RankUp (2024) & 5.74  & 8.82 & 5.89 & 4.22 \\
GCLSS (2025)  & 5.75 & 8.87 & 6.31 & 4.11 \\
DKD (ours)    & 5.77  & 8.90 & 5.91 & 4.17 \\
\bottomrule
\end{tabular}
}
\end{table}


\noindent\textbf{Running time analysis.}
Table~\ref{tab:inference} shows that DKD has inference latency comparable to RankUp and GCLSS on all four datasets. On UTKFace, DKD requires 5.77\,ms per sample, close to RankUp (5.74\,ms) and GCLSS (5.75\,ms). The same pattern holds on BVCC, Yelp, and MIMIC. In contrast, UCVME is consistently slower, which is consistent with its multi-pass inference procedure. These results indicate that DKD shifts its extra cost to training rather than deployment.

\noindent\textbf{GPU memory usage.}
The teacher network increases parameters during training, but in our
implementation, the peak GPU memory overhead of DKD over a single-model
baseline is $<\!1.4\times$.  Two practical factors contribute:
(1)~the teacher's unlabeled outputs are always detached before
entering any loss, so intermediate activations of that branch need
not be stored for gradient computation;
(2)~the prediction head is lightweight ($L\!\times\!M$) compared with
the backbone.
At inference time, only the student is loaded, so DKD's memory usage
is identical to a single-model baseline of the same backbone.



\section{Discussion}
Despite its effectiveness, DKD still has two limitations. First, DKD relies on a relatively large labeled set to train a sufficiently informative teacher model. When only extremely few labels are available ($<\!50$ on the evaluated datasets), the teacher predictions become unreliable, causing the variance-based weights to collapse toward nearly uniform values. A possible remedy is to introduce a warm-up stage that combines teacher guidance with a direct-regression prior. Second, our current theoretical analysis relies on the bounded teacher-noise assumption (Assumption~\ref{assum:noise}) and does not explicitly characterize pathological long-tail cases where a few samples exhibit extremely large $\sigma^{\mathcal T}$. Extending DKD with density-aware bin widths and robust clipping of $\omega_i$ would be a promising direction for improving its robustness in such regimes.

\section{Conclusion}

In this paper, we propose DKD, the very first dual-stream
knowledge distillation framework for fully exploiting the
abundance of unlabeled samples and noise mitigation in
pseudo-label supervision in SSR. DKD is designed to distill
both continuous-valued knowledge and distribution information, which better preserves regression magnitudes and enhances the sample efficiency. Then, the DDA is proposed
to further enhance the student’s capacity to mitigate noise
in pseudo-label supervision and learn a more well-calibrated
regression predictor. Extensive experimental results across
multiple datasets spanning four domains demonstrate that
DKD outperforms state-of-the-art methods, indicating its superior generalization capability.


\bibliographystyle{ACM-Reference-Format}
\bibliography{references}


\end{document}